\documentclass[11pt,a4paper]{article}

\usepackage[hyperref]{emnlp2020}
\usepackage{times}
\usepackage{latexsym}
\usepackage{graphicx}
\usepackage{tipa}
\usepackage{adjustbox}
\usepackage{tabularx}

\usepackage{multirow}
\usepackage{booktabs}
\usepackage{comment}
\usepackage{textcomp}
\usepackage{silence}
\usepackage{ulem} 
\usepackage[show]{boxnotes}

\newcommand{\eat}[1]{}

\iftrue 

  \renewcommand\sout[1]{\xspace}
\fi

\usepackage{microtype}

\aclfinalcopy 

\title{Translating Similar Languages: Role of Mutual Intelligibility in Multilingual Transformers}

\author{Ife Adebara ~~~~~~~~ El Moatez Billah Nagoudi ~~~~~~~~ Muhammad Abdul Mageed \\
\normalsize Natural Language Processing Lab  \\
  \normalsize The University of British Columbia\\
      
  \texttt{ \small \{ife.adebara,moatez.nagoudi,muhammad.mageed\}@ubc.ca}
  }

\date{}

\begin{document}

\maketitle
\begin{abstract}
We investigate different approaches to translate between similar languages under low resource conditions, as part of our contribution to the WMT 2020 Similar Languages Translation Shared Task. We submitted Transformer-based bilingual and multilingual systems for all language pairs, in the two directions. We also leverage back-translation for one of the language pairs, acquiring an improvement of more than 3 BLEU points. We interpret our results in light of the degree of mutual intelligibility (based on Jaccard similarity) between each pair, finding a positive correlation between mutual intelligibility  and model performance. Our Spanish-Catalan model has the best performance of all the five language pairs. Except for the case of Hindi-Marathi, our bilingual models achieve better performance than the multilingual models on all pairs.

\end{abstract}

\section{Introduction}
We present our findings from our participation in the WMT 2020 Similar Language Translation shared task, which
focused on translation between similar language
pairs in low-resource settings. Similar languages share a certain level of mutual intelligibility that may aid the improvement of translation quality. Depending on the level of closeness, certain languages may share similar orthography, lexical, syntactic, and or semantic structures which may make translation more accurate.  

The level of mutual intelligibility is such that speakers of one language can understand another language without prior instruction in that other language. They can also communicate without the use of a lingua franca which is a link or vehicular language used for communicating between speakers of different languages~\citep{gooskens2007contribution}. It is important to mention that, sometimes, the level of intelligibility varies in both directions. For instance, Slovene - Croatian intelligibility is said to be asymmetric such that speakers or Slovene can understand spoken and written Croatian better than speakers of Croatian understand Slovene ~\citep{golubovic2015mutual}. 

Machine translation of similar languages has been explored in a number of works~\citep{hajic2000machine, currey2016using,dabre2017empirical}. This can be seen as part of a growing need to develop models that translate well in low resource scenarios. The goal of the current shared task is to encourage researchers to explore methods for translating between similar languages. We also view the shared task as useful context for studying interaction between degrees of similarity and mutual intelligibility on the one hand, and model performance on the other hand.  We explore the use of bilingual and multilingual models for all the 5 shared task language pairs. We also perform back-translation for one language pair.
 
In the remainder of this paper, we discuss related literature in Section \ref{Lit-review}. We explain the methodology which includes a description of the Transformer model, back-translation and beam search in Section \ref{sec:Methodology}. In Section \ref{Models}, we describe the models we developed for this task and we discuss the various experiments we perform. We also describe the architectures of the models we developed. 
Then we discuss the evaluation procedure in Section \ref{Evaluation}. Evaluation is done on both the validation and test sets. We conclude with discussion of the insights we gained from the shared task in Section \ref{Conclusion}. 

\section{Related Work}\label{Lit-review}
Translation between similar languages has recently attracted attention. Different approaches have been adopted using state-of-the-art techniques, methods, and tools to take advantage of the similarity between languages even in low resource scenarios. Approaches that have been effective for other machine translation tasks have proven to achieve success in the context of similar language translation as well.

NMT models, specifically the Transformer architecture, has been shown to perform well when translating between similar languages \citep{baquero2019mllp,przystupa2019neural}. The use of in-domain data for fine-tuning has also proven to be of remarkable benefit for this task. This problem has also been tackled both by using character replacement to leverage the orthographic and phonological relationship between closely related mutually intelligible language pairs~\citep{chen2019machine}. A new approach was also introduced for this task using a two-dimensional method that assumes that each word of the target sentence can be explained by all the words in the source sentence~\citep{baquero2019mllp}.

Within the realm of MT for low resource languages, recent work has focused on translation using large monolingual corpora due to the scarcity of parallel data for many language pairs ~\citep{lample2018phrase,lample2017unsupervised,artetxe2017unsupervised}. These approaches have leveraged careful initialization of the unsupervised neural MT model using an inferred bilingual dictionary, sequence-to-sequence language models, and back-translation to achieve remarkable results. The bilingual dictionary is built without parallel data by using an unsupervised approach to align the monolingual word embedding spaces from each language ~\citep{conneau2017word,artetxe2018generalizing}. Since parallel data is not available in sufficiently large quantities, back-translation is used to create pseudo parallel data. The monolingual data of the target language is translated into the source using an existing translation system (e.g., one trained with available gold data). The output is then used to train a new MT model ~\citep{sennrich2015improving}. Weak supervision caused by back-translation results in a noisy training dataset. This eventually can affect translation quality.

More recent works adopt different approaches to manage noise in back-translation. For instance, phrase based statistical MT models are introduced as a posterior regularization during the back-translation process to reduce the noise and errors of the data generated ~\citep{ren2019unsupervised}. Another method ~\citep{artetxe2019effective} uses cross lingual word embeddings incorporated with sub-word information. The weights of the log-linear model is then tuned through an unsupervised process and the entire system is jointly refined in opposite directions to improve performance. This method out-performs previous SOTA model with about 5-7 BLUE points. A re-scoring mechanism that re-uses the pre-trained language model to select translations generated through beam search has also been found to improve fluency and consistency of translations ~\citep{liu2019incorporating}. Yet another approach, combines cross-lingual embeddings with a language model to make a phrase-table ~\citep{artetxe2019bilingual}. The resulting system is then used to generate a pseudo parallel corpus with which a bilingual lexicon is derived. This approach can work with any word or cross-lingual embeddings techniques. 

\section{Methodology}\label{sec:Methodology}
Motivated by the success of Transformers and back-translation, we develop a sequence-to-sequence approach using the Transformer architecture. We also perform back-translation for one language pair. For decoding, we use Beam Search (BS). BS is an heuristic decoding strategy based on exploring the solution space and selecting a sequence of words that maximize the overall likelihood of the target sentence. During the translation, we hold a beam of $\beta$ sequences  (\textit{beam size}) which are iteratively extended. At each step, $\beta$ words are selected to extend each of the sequences in the beam, so the output is $\beta^{2}$ candidate sequences (hypotheses), we retain only the $\beta$ highest score hypotheses for the next step (top-$\beta$ candidates) \cite{koehn2009statistical}. In all our experiments we use beam size of $5$ whilst decoding. 

\subsection{Transformer}
Our baseline models are based on the Transformer architecture. A Transformer  ~\citep{vaswani2017attention} is a sequence-to-sequence model that does not have the recurrent architecture present in Recurrent Neural Networks (RNNs). It uses a positional encoding that can remember how sequences are fed into the model. These positions are added to the embedded representation (n-dimensional vector) of each word. Transformers have been shown to train faster than RNNs for translation tasks. 

The encoder and decoder in a Transformer model have modules that consist mainly of Multi-Head Attention multi-head attention and Feed Forward feedforward layers. The attention mechanism is based on a function that  operates on Q (\textit{queries}), K (\textit{keys}), and V (\textit{values}). The query is a vector representation of one token in the input sequence, K refers to the vector representations of all the tokens in the input sequence. More information about the Transformer are in ~\citep{vaswani2017attention}.

\subsection{Back-translation }
We perform back-translation using the monolingual model developed for the Croatian-Slovene (HR-SL) language pair. We use the best HR-SL model checkpoint that acquire the highest BLEU score on the DEV set to translate the monolingual HR data. This produces synthetic Slovene (SL) data which we then use as the source language while the original monolingual data is used as target when training the SL-HR model. We combine this data with the initial training data. Due to time constraints, we used only a subset of the monolingual data with a beam size of 5. 

\subsection{Jaccard Similarity}
Jaccard similarity compares similarity, diversity, and distance of data sets \cite{niwattanakul2013using}. It is calculated between two data sets in our case two languages) A and B by dividing the number of features common to the two sets (their intersection) by the union of features in the two sets, as in (1) below: 

\begin{equation}
J(A,B) = \frac{|A\cap B|}{| A \cup B|}
\end{equation}  

We use tokens, identified based on white space, as features when we calculate Jaccard. 

\section{Experiments} \label{Models}

\subsection{Model Architecture}
Our neural network models are based on the Transformer architecture (as described in Section \ref{sec:Methodology}) implemented by Facebook in the Fairseq toolkit. The following hyper-parameter configuration was used: $6$ attention layers in the encoder and the decoder, $4$ attention heads in each layer, embedding dimension of $512$, maximum number of tokens per batch was set to $4,096$, Adam optimizer with $\beta{1} = 0.90$, $\beta{2} = 0.98$, dropout regularization was set to $0.3$, weight-decay was set at $0.0001$, $label$-$smoothing = 0.1$, variable learning rate set at $5e^{-4}$ with the inverse square root, lr-scheduler and $warmup$-$updates = 4,000$ steps. We used the label smoothed cross-entropy criterion, and gradient clip-norm threshold was set to $0$. 


\section{Data} \label{Data}
We used all the parallel data for all language pairs \url{http://www.statmt.org/wmt20/similar.html}. The task was constrained so we did not add any additional data to develop our models. We used the monolingual data for the SL-HR language pair for back-translation. Table \ref{tab:data}  shows the size of the data in terms of the number of sentences and words for each language pair while Table \ref{tab:exp} shows example source and corresponding outputs from our bilingual and multilingual models for each language pair. We also calculated the jaccard similarity for the training data we used for the tasks. ``Jaccard similarity" measures the similarity between two text documents by taking the intersection of both and dividing it by their union. Linguists measure these intersections \citep{oktavia2019understanding, gooskens2017linguistic} between languages to determine the level of mutual intelligibility as well as classify languages as dialects of the same language or different languages. We calculated Jaccard similarity for each language pair.

\begin{table*}[t]
\small
\centering
\begin{adjustbox}{width=16cm}
\renewcommand{\arraystretch}{1.5}
{
        \begin{tabular}{>{}clll}
        \toprule

      \textbf { \small Model }    &\textbf { \small Pair }   &\textbf {\small Sentence } & \textbf{ \small Translation} \\    \toprule

 \multicolumn{1}{c}{}  & \multirow{2}{*}{\textbf{Es-Ca}   } & Diseña stickers para soñar   &      \small{ Dissenya stickers per soyir }\\ 
 \multicolumn{1}{c}{}  &     & Mueva el diez de corazones al nueve de corazones .  &      \small{ Mobles el deu de garrons al nou de garrons . }\\ 
\cline{2-4}

 \multicolumn{1}{c}{}  & \multirow{2}{*}{\textbf{Es-Pt}   } & Diseña stickers para soñar     &      \small{  Design stickers para sonhar}\\ 
 \multicolumn{1}{c}{}  &     &Mueva el diez de corazones al nueve de corazones .  &      \small{ Muda o dez corações para nove corações .}\\ 
\cline{2-4}

 \multicolumn{1}{c}{}  &   \multirow{2}{*}{\textbf{Sl-Hr}   }   &Vesel pomladni pozdrav ob novi izdaji Bisnode novičk .    &      \small{ Sretan proljetni pozdrav na novom izdanju Bisnode Vijesti .}\\ 
 \multicolumn{1}{c}{\multirow{-4}{*}{\rotatebox[origin=c]{90}{\textbf{\small Bilingual  Model }}}}  &    & Z lepimi pozdravi ,  &      \small{  S lijepim pozdravima ,}\\ 

\cline{2-4}
    
 \multicolumn{1}{c}{}  &  \multirow{2}{*}{\textbf{Sl-Sr}   }&Danes ni enostavno slediti vsem informacijam , ki so pomembne za poslovanje .      &      \small{  Danas nije lako pratiti sve informacije koje su važne za poslovanje . }\\ 
 \multicolumn{1}{c}{}  &     & Iščete podatke za drugo državo ?   &      \small{ Tražite podatke za drugu zemlju ?}\\ 
\cline{2-4}

    \toprule 
 
 \multicolumn{1}{c}{}  & \multirow{2}{*}{\textbf{Es-Ca}   } & Mueva el diez de corazones al nueve de corazones .     &      \small{ Muva el 10 de coração al 9 de coraons .}\\ 
 \multicolumn{1}{c}{}  &     &  el cuatro de diamantes        &      \small{el quatre de diamants }\\ 
\cline{2-4}

 \multicolumn{1}{c}{}  & \multirow{2}{*}{\textbf{Es-Pt}   } & Luche en el aire con un avión enemigo         &      \small{Luche en l aire amb un avió enemigo   }\\ 
 \multicolumn{1}{c}{}  &     &  Entonces , ¿ qué salió mal ?    &      \small{Então , o que saiu mal ?   }\\ 
\cline{2-4}
    
 \multicolumn{1}{c}{}  &   \multirow{2}{*}{\textbf{Sl-Hr}   }   &  Objašnjenje - Indeks plaćanja Datum :      &      \small{  Objašnjenje -- Indeks plaćanja Datum : }\\ 
 \multicolumn{1}{c}{\multirow{-4}{*}{\rotatebox[origin=c]{90}{\textbf{\small Multilingual  Model }}}}  &    & Poštovani ,  &      \small{Poštovani ,   }\\ 

\cline{2-4}
    
 \multicolumn{1}{c}{}  &  \multirow{2}{*}{\textbf{Sl-Sr}   }& Iščete podatke za drugo državo ?    &      \small{ Tražite podatke za drugu zemlju ?  }\\ 
 \multicolumn{1}{c}{}  &     & Vesel pomladni pozdrav ob novi izdaji Bisnode novičk .  &      \small{ Sretan proljetni pozdrav uz novo izdanje Bisnode novosti .  }\\ 
\cline{2-4}

    \toprule

\end{tabular}}
\end{adjustbox}
    \caption{Examples sentences from the various pairs and corresponding translations based on the bilingual and multilingual models. Examples are from the DEV set. }
    
    \label{tab:exp}
\end{table*}

\begin{table}[!ht]
\begin{center}
\textbf{}
\small 
\begin{tabular}{>{}clll}
 \hline
\multicolumn{1}{c}{}  &\textbf{Language} & \textbf{\#Sentences} & \textbf{\#Words} \\ 
 \hline

\multicolumn{1}{c}{}  &hi & $43.2K$ & $829.9K$ \\

\multicolumn{1}{c}{}  &mr & $43.2K$ & ~$600K$  \\

\multicolumn{1}{c}{\multirow{-2}{*}{\rotatebox[origin=c]{90}{\textbf{\small Hi-Mr}}}}  &mono-hi & $113.5M$ & $4.74B$  \\

\multicolumn{1}{c}{}   &mono-mr & $4.9M$ & $112.6M$ \\

 \hline
 
 \multicolumn{1}{c}{}  &es & $11.3M$ & $150.4M$ \\

\multicolumn{1}{c}{}  &ca & $11.3M$ & $163M$  \\

\multicolumn{1}{c}{\multirow{-2}{*}{\rotatebox[origin=c]{90}{\textbf{\small Es-Ca}}}}  &mono-es & $58.4M$ & $1.5B$  \\

\multicolumn{1}{c}{}   &mono-ca & $28M$ & $763.7M$ \\

 \hline

 \multicolumn{1}{c}{}  & es & $4.15M$ & $86.6M$ \\
\multicolumn{1}{c}{}  & pt & $4.15M$  &  $82.5M$   \\

\multicolumn{1}{c}{\multirow{-2}{*}{\rotatebox[origin=c]{90}{\textbf{\small Es-pt }}}}  &mono-es &  $58.4M$   &  $1.47B$  \\

 \multicolumn{1}{c}{}  &mono-pt &  $11.4M$ & $233.9M$\\

\hline

\multicolumn{1}{c}{}  &sl & $17.6M$ & $113.09M$ \\

\multicolumn{1}{c}{}  &hr &  $17.6M$ & $117.73M$   \\

\multicolumn{1}{c}{\multirow{-2}{*}{\rotatebox[origin=c]{90}{\textbf{\small Sl-Hr}}}}  &mono-sl & $46.25M$ & $770.6M$  \\

\multicolumn{1}{c}{}   &mono-hr & $64.5M$ & $1.24B$ \\

\hline 

\multicolumn{1}{c}{}  &sl &  $14.1M$ & $79.1M$   \\

\multicolumn{1}{c}{}  &sr &  $14.1M$ & $86.1M$   \\

\multicolumn{1}{c}{\multirow{-2}{*}{\rotatebox[origin=c]{90}{\textbf{\small Sl-Sr}}}}  &mono-sl & $46.2M$ & $770.6M$  \\

\multicolumn{1}{c}{}   &mono-sr & $24M$ & $489.9M$ \\
 \toprule

\end{tabular}
\end{center}
\caption{Number of sentences and words for the training data used for each language pair.}\label{tab:data}
\end{table}

\subsection{Pre-processing}
Pre-processing was by a regular Moses toolkit ~\citep{koehn2007moses} pipeline that involved tokenization, byte pair encoding  and removing long sentences. We applied Byte-Pair Encoding (BPE) ~\citep{sennrich2015neural} operations, learned jointly over the source and target languages. For each language pair, we used $32$k split operation for subword segmentation \citep{sennrich-etal-2016-neural}. We run experiments with Transformers under three settings, as we explain next. 

\subsection{Models}
We develop both bilingual and multilingual models using gold data for all pairs. For one pair, we also use back-translation with one bilingual model. We provide more details next.
\subsubsection{Bilingual Models} 
In this setting, we  build an independent model for each language pair. We develop models for both directions for all language pairs, thus ultimately creating 12 models (6 for each direction). We train each model on 1 GPU for 7 days. 


\subsubsection{Multilingual Models} 
We develop two multilingual models that translates between all languages; a model for each direction (2 models overall) \cite{johnson2017google}. We add a language code representing the target language as the start token for each line of the source data. We train each model on 4 GPUs for 7 days. We use the same hyper-parameters values set for the bilingual models. Multilingual models enable us to determine the impact of learning similar languages with a shared representation.

\subsubsection{Bilingual Model with Back-translation }
For the third approach, we combine back-translation with the bilingual translation model for the SL-HR language pair. We incorporated the monolingual data to do this. This was influenced by report ~\citep{sennrich-etal-2016-improving} in literature on the significant improvement of translation quality when monolingual data is incorporated into training data through back-translation. We were able to test the effect of back-translation on one model.

\section{Evaluation} \label{Evaluation}
We evaluated both the DEV and TEST sets. We used the best-checkpoint metric with BLEU score to evaluate the validation set at each iteration. We used a beam size of 5 during the evaluation. We de-tokenized from BPEs back into words.

\subsection{Evaluation on DEV set}
We report the results on the DEV sets for each language pair in Table \ref{tab:bleu}. These models were trained without the monolingual data except the SL-HR pair with the asteriks. 
\begin{table}[ht]
\begin{center}
\small 
\begin{tabular}{p{1cm}p{2.5cm}p{2.5cm}} 
 \toprule
\textbf{Pair} & \textbf{Bilingual Models} & \textbf{Multiling. models} \\ 
 \toprule
hi-mr & 12.14 & 16.35 \\
mr-hi & 10.63 & 01.02  \\
es-ca & 74.85 & 16.13 \\
ca-es &  74.24 & 64.57  \\
es-pt & 46.71 & 26.41 \\
pt-es & 41.12 &  05.81 \\
$*$sl-hr & 36.89 & - \\ 
sl-hr & 33.28 & 09.25 \\
hr-sl & 55.51 &  07.94 \\
sl-sr & 40.80 & 32.80  \\
sr-sl & 39.80 & 06.97 \\
 \toprule
\end{tabular}
\end{center}
\caption{Evaluation in BLEU on the development set for the different language pairs. The asteriks shows the model with back-translation. }
\label{tab:bleu}
\end{table}

The bilingual models outperform the multi-lingual models for all language pairs except the hi-mr language pair. 
\subsection{Evaluation on TEST}
In order to evaluate the test data, we removed the byte-pair code from the test set. We used the fairseq-generate mode while translating the test set. We show results on the test set in Table \ref{tab:bleu-2}. 

\begin{table}[ht]
\begin{center}
\small 
\begin{tabular}{p{1cm}p{2.5cm}p{2.5cm}} 
 \toprule
\textbf{Pair} & \textbf{Bilingual Models} & \textbf{Multiling. models}  \\ 
 \toprule
hi-mr & 0.49 & - \\

es-ca & 41.74 & 8.49 \\

ca-es &  45.23 & 45.86 \\

es-pt & 23.35 & 17.06 \\

pt-es & 24.26 &  21.55 \\

*sl-hr & 20.92 & 22.26  \\

hr-sl & 14.94 &  7.37 \\

sl-sr & 14.7 & 20.18  \\

sr-sl & 19.46 & 11.37 \\
 \toprule
\end{tabular}
\end{center}
\caption{The BLEU scores for some of the models on the test set. The language pair with the asterisks has back-translation. \footnote{We did not include results for some models because we did not receive the results before we submitted this paper.} }\label{tab:bleu-2}
\end{table}

\subsection{Discussion}
\begin{figure}[ht]
\centering
  \includegraphics[width=8cm,height=4.5cm]{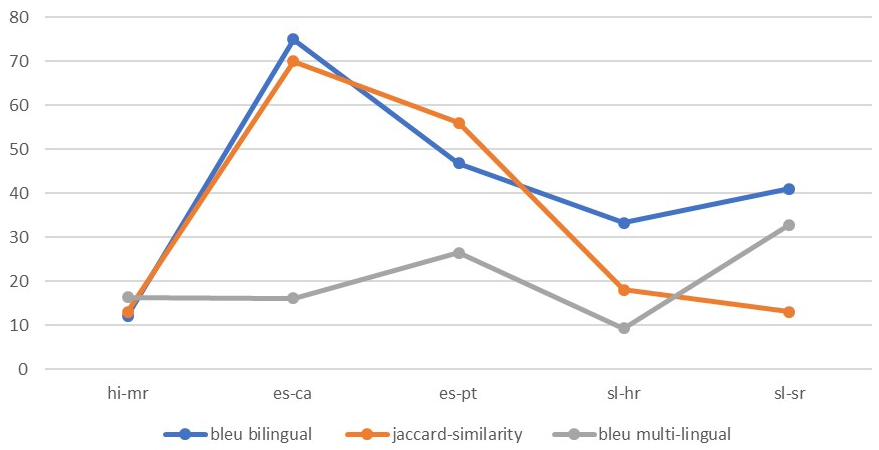}
  \caption{Interaction between performance in BLEU and Jaccard similarity.}
\label{fig:our_map}
\end{figure}

We used the Jaccard similarity to measure the level of mutual intelligbility. Figure ~\ref{fig:our_map} shows a positive correlation between the BLEU scores and the Jaccard similarity between each language pair.~\footnote{We multiplied the jaccard similarity by 100 to reduce the range of values on the y axis. } This relationship hold both for the bilingual and multilingual models. One exception is the Slovene-Serbian pair (SL-SR) where higher similarity does not translate into higher BLEU.
For example, the SL-SR BLEU is below the SL-HR BLEU even though the latter pair has a higher similarity score. Interpreting Jaccard to mean mutual intelligibility, our findings imply a higher intelligibility is correlated with higher BLEU scores. However, there is a need to further investigate this relationship due to the SL-SR we observe.

\section{Conclusion} \label{Conclusion}

We described our contribution to the WMT2020  Similar Languages Translation Shared Task. We developed both bilingual and multilingual models for all pairs, in both directions. We showed back-translation to help improve performance on one pair. We also showed how mutual intelligibility between a pair of languages ( measured by Jaccard similarity) positively correlates with model performance (in BLEU). Future work can focus on exploiting other similarity metrics and providing a more in-depth study of mutual intelligibility between similar languages and how it interacts with MT model performance both in bilingual and multilingual models. The utility of back-translation on pairs we have not studied can also be fruitful.

 \section*{Acknowledgements}
 We gratefully acknowledge support from the Natural Sciences and Engineering Research Council of Canada (NSERC), the Social Sciences Research Council of Canada (SSHRC), Compute Canada (\url{www.computecanada.ca}), and UBC ARC--Sockeye (\url{https://doi.org/10.14288/SOCKEYE}).

\bibliography{anthology,emnlp2020}
\bibliographystyle{acl_natbib}

\end{document}